# ATM Fraud Detection using Streaming Data Analytics

**Yelleti Vivek[1], Vadlamani Ravi[1]•, Abhay Anand Mane[2], Laveti Ramesh Naidu[2]**

[1]Centre for Artificial Intelligence and Machine Learning
Institute for Development and Research in Banking Technology,
Castle Hills Road #1, Masab Tank, Hyderabad-500076, India
[2] Center for Development of Advanced Computing,
#1 Old Madras Road, Bangalore, Karnataka 560038, India
yvivek@idrbt.ac.in ; vravi@idrbt.ac.in; abhaym@cdac.in; rameshl@cdac.in

## Abstract

Gaining the trust and confidence of customers is the essence of the growth and success of financial institutions and organizations. Of late, the financial industry is significantly impacted by numerous instances of fraudulent activities. Further, owing to the generation of large voluminous datasets, it is highly essential that underlying framework is scalable and meet real time needs. To address this issue, in the study, we proposed ATM fraud detection in static and streaming contexts respectively. In the static context, we investigated a parallel and scalable machine learning algorithms for ATM fraud detection that is built on Spark and trained with a variety of machine learning (ML) models including Naive Bayes (NB), Logistic Regression (LR), Support Vector Machine (SVM), Decision Tree (DT), Random Forest (RF), Gradient Boosting Tree (GBT), and Multi-layer perceptron (MLP). We also employed several balancing techniques like Synthetic Minority Oversampling Technique (SMOTE) and its variants, Generative Adversarial Networks (GAN), to address the rarity in the dataset. In addition, we proposed a streaming based ATM fraud detection in the streaming context. Our sliding window based method collects ATM transactions that are performed within a specified time interval and then utilizes to train several ML models, including NB, RF, DT, and K-Nearest Neighbour (KNN). We selected these models based on their less model complexity and quicker response time. In both contexts, RF turned out to be the best model. RF obtained the best mean AUC of 0.975 in the static context and mean AUC of 0.910 in the streaming context. RF is also empirically proven to be statistically significant than the next-best performing models.

## 1. Introduction

Customers ubiquitously utilize automated teller machines (ATM) as one of the widely accepted forms of payment on a "*day-to-day*" basis. Over the past two decades, the demand for ATMs has surged as financial institutions strive to fulfil client demand and provide sophisticated service management. However, this immense growth attracted the scammers attention, who have targeted and caused significant financial loss. This inflicts a severe loss to the customers and further affects the '*bank-customer*' relationship. This could be avoidable by taking proper measures which indeed gain the trust and builds the confidence of the customer over the financial institutions and organizations.

---

* Corresponding Author

There are different sorts of financial fraud exist, including credit card fraud, money laundering, debit card fraud etc. When any of these financial crimes occur, financial organisations suffer significant losses. Therefore, it is imperative for any financial institution or organisation to handle them carefully. The negative repercussions of various financial frauds have captivated the consciousness of stakeholders. Although there is no universally accepted definition of financial fraud, Oxford English Dictionary[1] defines fraud as "*wrongful or criminal deception intended to result in financial or personal gain.*" Phua et al. [2] describe fraud as "*abusing a profit organization's system without necessarily leading to direct legal consequences*". Wang et al. [14] define it as "*a deliberate act that is contrary to law, rule, or policy with the intent to obtain an unauthorized financial benefit.*" Currently, the prosperity of the financial institution and the customer relationship management is entailed with these financial instruments.

Organizations base their policies and critical decisions on the analyses of the vast volumes of data, or "Big Data". These financial fraud datasets in particular fall within the big data paradigm. The five Vs – Volume, Velocity, Variety, Veracity, and Value [11] are the primary characteristics of big data. There are several big data processing tools, such as Hadoop and Spark, which enable the organizations to collect and manage huge data. Recently, Spark, has gained popularity owing to its unified data analytics engine which can handle massive datasets. Spark offers a variety of abstractions, including data frames (DF) and resilient distributed data frames (RDD), which are immutable, support persistence, are cacheable, and are partitioned into multiple data partitions thereby supporting parallel and distributed operations. Additionally, Spark offers extensions for various programming languages like Scala, Python, Java, and R. Further, Spark extends its support to SQL with SparkSQL, developing scalable machine learning algorithms with SparkMLlib, and streaming analytics with Discretized streams (DStreams) as the abstraction. DStream, is a collection of RDDs collected over time that facilitates batch stream processing. Spark can easily integrate with other ingestion tools such as Apache Kafka, Kinesis, or TCP sockets and other big data processing tools such as Hadoop, Hive, etc. making it to manage the continuous stream of data.

Among the types of financial frauds discussed earlier, ATM fraud detection has captivated our interest. The ATM transaction history can be examined to detect fraudulent transactions. We determined that the following scenarios might assist us in identifying fraudulent transactions: (i) stolen or lost debit/credit card, (ii) unusual transaction history, (iii) transactions initiated from an unusual location, and (iv) transaction involving bulk amounts, which could increase the probability of becoming fraudulent. It is quite difficult to carry out such analysis in real-time within limited time interval. Therefore, there is a need to automate this process by employing various machine learning techniques. Detecting ATM fraud detection is indeed a binary classification task, where fraudulent transactions being are regarded as a positive class and non-fraudulent transactions as a negative class.

Owing to the significance of identifying fraud detection under the big data paradigm. Through this study, we addressed the ATM fraud detection in the big data paradigm using a scalable machine learning algorithms and efficient solution methodology in both static and streaming contexts. To achieve this, we conducted the study as follows:

- Firstly, in the static context, we developed machine learning models on the historic ATM transactions data collected from India. To create scalable ML models, we used Spark MLlib.
- Secondly, in the streaming context, we used DStreams to capture data at regular intervals and employed a sliding window based fraud detection approach.

---

[1] Oxford Concise English Dictionary, Tenth ed, Publisher, 1999

In the current study, we addressed the following challenges in both static and streaming contexts as follows: (i) handling imbalance in the fraud detection datasets, (ii) developing scalable ML models to meet the requirements of the big data paradigm, (iii) in the streaming context, the latency should be very less and time-time to analysis need to be provided without compromising on the performance, (iv) the streaming methodology should support a large number of data streams, and (v) automating the entire fraud detection process under big data paradigm.

In the current study, we analyzed an India-sourced ATM transactions data for fraud detection. As discussed earlier, this ATM fraud detection problem has a significant impact on daily living has motivated us to work towards providing and effective and efficient solution. In this study, we performed ATM fraud detection using binary classification tasks. To accomplish this, we employed several machine learning techniques: NB, LR, SVM, DT, RF, GBT, and MLP. The collected ATM transactions data is highly unbalanced in nature. This data imbalance is handled by employing the following techniques such as SMOTE and its variants (SMOTE-ENN, SMOTE-Tomek), and ADASYN. Further, we employed the GAN variants (i.e., V-GAN and W-GAN) too owing to its recent advancements and achievements.

The major contributions of the proposed work are as follows:

- Proposed the resilient ATM fraud detection framework under Spark.
- Explored and compared ML algorithms for ATM fraud detection in the static context.
- Investigated the use of GAN variants (V-GAN and W-GAN) for handling data imabalance in ATM fraud detection.
- Proposed sliding window approach based ATM fraud detection using streaming analytics thereby handling the latency and time-to-time decision.
- Handled the imbalance by employing various data balancing techniques.

The remainder of the paper is arranged as follows: Section 2 presents the literature review. Section 3 discusses the proposed framework for ATM transactional data using binary modeling under static and streaming contexts. Section 4 presents the background theory of machine learning and the balancing techniques employed. Section 5 describes the dataset. Section 6 discusses the results of various techniques. Finally, Section 7 concludes the paper.

## 2. Literature Review

This section will cover pertinent research that has been used to mitigate fraud detection in the banking and financial industries related to credit card fraud detection, financial statements, etc. in both static and streaming contexts respectively.

Data mining algorithms play a quintessential role in detecting fraud post-facto [1]. These consist of LR, MLP, DT, etc. When cutting-edge methods, such as semi-supervised and unsupervised techniques, are used to mine the data that arrives in a stream, it may also be discovered in almost real-time [2,3]. Online classification, online clustering, outlier detection, etc., are some of the most sophisticated approaches. A reasonable amount of research is reported concerning credit card fraud detection [1,4-5], however, ATM/Debit card fraud detection is seldom the researched topic. To the best of our knowledge, one research article appeared dealing with ATM transactional frauds, where a specialized language is proposed for proactive fraud management in financial data streams [3]. Depending, on how an ATM card is used – at an ATM, a point of sale (POS) device, or online banking, there are three alternative options. Other frauds can be perpetrated, and as the resultant data varies from structured to semi-structured to unstructured, so does the method they can be detected. Wei et al. [15] observed various characteristics and challenges involved in the credit card fraud detection. The development of a scalable system for credit card fraud detection utilizing distributed settings was pioneered by Chan et al. [6]. The authors demonstrated the significance of considering scalability into account when developing the solution.

Chen et al. [7] proposed a binary support vector machine, a hybrid of SVM and genetic algorithm (GA). Here, the optimal mix of support vectors is chosen using a genetic algorithm. In order to estimate the distribution of the input data, self-organizing map (SOM) is also included. In order to fulfil real-time analytics, Dorronsoro et al. [8] proposed an online system for credit card fraud detection based on the neural classifier. To detect fraud, the authors proposed a non-linear variation of fisher's discriminant analysis. This strategy could separate a significant amount of fraudulent activities from legitimate traffic. A real-time fraud detection system was presented by Quah et al. [9] that uses the SOM to identify fraudulent credit card transactions and consumer behaviour online. Sanchez et al. [10] applied association rule mining to find the pattern of anomalous behaviour that would assist them to identify a fraudulent transaction. On the transaction database, the authors showed how well the suggested technique worked. Hidden markov model (HMM), Srivastava et al. [11] devised a method that comprises the following for credit card fraud detection: (i) Initially, the set of customers with identical spending patterns are clustered together by using clustering technique. (ii) After that, HMM is executed to identify the critical patterns and sequences which capture the fraudulent transactions. Zaslavsky and Strizhak [12] applied the SOMs for credit card fraud detection. IT-based paradigm was presented by Wang et al. [14] to address a variety of frauds: (i) system attacks, and (ii) non-system attacks. The authors studied the effectiveness of their framework on Taiwan financial fraud dataset.

Credit card transaction data frequently concern scalability as one of the primary concerns that must be resolved in real-time. Syeda et al. [13] developed a parallel granular neural network with shared memory architecture to address the above mentioned issue. It is applied to visa card transactions to identify fraudulent transactions. Wei et al. [15] used the transaction vector to identify and store customer behaviour to handle this imbalance issue of fraudulent datasets. The authors proposed a contrast miner, that efficiently extracts critical information by mining the contrast patterns and identifying the unlawful behaviour. Due to the considerable increase in online activities, several fraudsters are attacking the most often used websites. There are a number of efforts proposed to identify such attacks in the literature [31]. Arya et al. [16] applied a spiking neural network is inspired by neuroscience, and applied it to several classification problems.

Ravisankar et al. [27] proposed a fraud detection framework to identify the critical financial ratios obtained by analyzing the financial statements. With this, it would be possible to assess whether the company is carrying out too much debt or inventory. The performance of the models is frequently hampered by the presence of missing values in the financial data. Fraquad et al. [29] proposed a modified active learning framework, which addresses churn prediction and insurance fraud detection, uses the SVM to extract the if-then rules. It is a three-phase approach where: (i) SVM is used for the recursive feature elimination, (ii) synthetic data is generated by using active learning, and (iii) rules are generated by using DT and NB. Vasu and Ravi [30] proposed an undersampling technique to address imbalance challenge in many banking and finance applications. K-Reverse nearest neighbourhood is used to remove the outliers from the majority class. Then, the classifiers are used to predict fraudulent transactions. Sundarkumar and Ravi [23] analyzed the dataset available in the CSMINING group. The authors invoked mutual information as the feature selection criteria are applied after extracting the features from the API calls. The dataset is imbalanced in nature. Hence, the authors employed oversampling techniques over several ML techniques.

Customers sometimes wonder if they are conducting transaction at the rightful website. Determining if the website is a target of a cyberattack is therefore critical. A strategy to recognise phishing websites was presented by Dhanalakshmi et al. [34]. To determine the validity of a website, their methodology examines the server's IP address, domain, and internet domains to identify website legitimacy. Lakshmi and Vijaya [33] proposed a supervised learning mechanism to accomplish phishing detection. The extraction the features from 200 website URLs by the authors served as evidence of the success of their methodology. PSO-trained classification association rule mining (PSOCARM) [18], a rule based classifier indicates whether or not phishing is detected. Further, Tayal and Ravi [19] developed a fuzzy rule-based association rule mining by formulating the above as a combinatorial optimization problem. This fuzzy-based rule miner is trained by using PSO and applied to a transactional database to extract the rule, which will help to detect phishing attacks. Pradeep et al. [22] also proposed a rule-based classifier based on firefly and threshold-accepting algorithms to classify fraudulent financial statements.

Malware detection poses a number of difficulties and needs a robust, advanced security system. Application program interface (API) calls are typically used by malware attackers to steal credit card details, important personal data, etc. In order to retrieve the information, Sundarkumar et al. [20] employed this API call information and topic modelling approach. Their findings demonstrate that DT and SVM outperformed the competition, with DT securing first place in the interpretability category. Another study by Sheen et al. [32], uses features retrieved from executable files and an ensemble of classifiers to identify malware. The authors' main goal was to employ the harmony search algorithm with the fewest possible classifiers. On other hand, fraud detection also can be solved by using One class classification (OCC) methods. Kamaruddin and Ravi [17] implemented the above hybrid architecture, i.e., PSO-AANN architecture, under the Spark environment to make it suitable for big datasets. The proposed architecture outperformed the extant approaches, such as OC-SVM, etc., The rarity problem in fraud detection is handled by, Sundarkumar and Ravi [21], who proposed an undersampling method where the K-Reverse nearest neighbourhood and OCSVM are applied in tandem and remove the unnecessary samples in the majority class. The authors demonstrated their approach effectiveness in the automobile insurance fraud detection dataset and customer credit card churn prediction dataset.

Now, we will review the studies related to fraud detection solving using real time analytics. Thennakoon et al. [46] utilized ML model and application processing interface (API) mechanism to flag the fraudulent transactions. Arya and Sastry [49] proposed a credit card fraud detection based on predictive modelling and named it deep ensemble learning for detecting fraudulent transaction in real-time data stream. The authors tried to address the overfitting issues which is one of the major concern of the previous studies. They employed ensemble of deep learning network to identify the credit card frauds. Mittal and Tyagi [48] discussed the pros and cons of various computational techniques for real time credit card fraud detection. Abakiram et al. [50] proposed a real time model for credit card fraud detection based on deep learning. The authors employed deep auto encoder to classify the fraudulent transactions. Rajeswari and Babu [45] proposed a credit card fraud detection system in the real time. The authors proposed a framework where the models are developed and relied on historical data with which the fraudulent transactions are identified in real-time. Carcillo et al. [47] proposed a scalable framework for credit card fraud detection using spark. The authors integrated the ingestion tools such as Kafka, Cassandra into their framework and simulated the credit card fraud transactions coming in the form of data streams. The employed three different mechanisms to identify the fraudulent transactions.

## 3. Proposed Methodology

This section starts with the proposed methodology in the static context and then continues with a discussion on the streaming context.

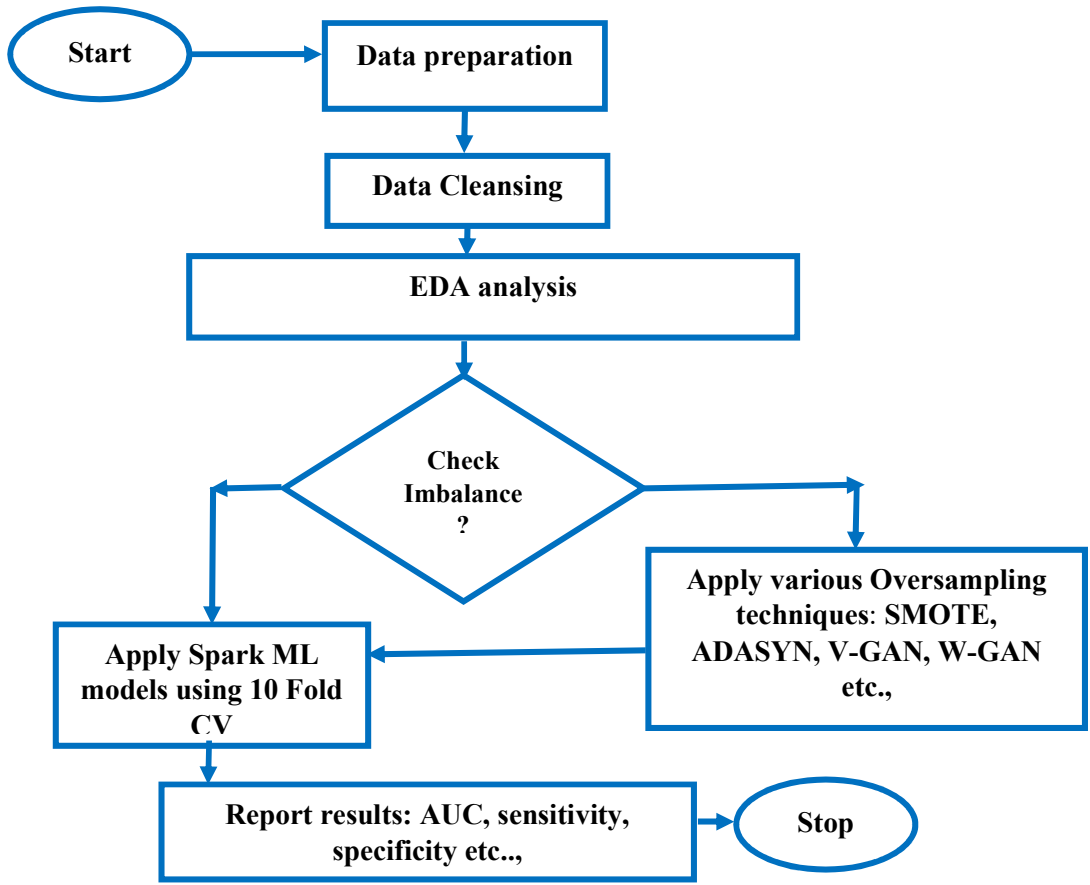

**Fig. 1. Flowchart of the proposed methodology in the static environment**

### 3.1 Binary Classification Methodology Under Spark

The proposed methodology is depicted in Fig. 1, and the block diagram is depicted in Fig. 2, and can be used in a static environment, where the model is trained on historical data. The process starts with the data collection step. In the current work, we collected ATM transaction data from India. In India, all the banks maintain software to monitor the daily transactions occurred in an ATM. The collected dataset contains transactions that occurred in three different modes viz., (i) ATM transaction data, (ii) at the point of sale (POS), and (iii) internet transaction. It includes critical private information related to the customers. To maintain the data integrity and privacy of the customers, we masked the feature names.

Since, 'garbage data in results in garbage analysis out' is very suitable in machine learning. Therefore, to increase the richness of the data, we include the following data cleansing steps: (i) removing duplicate transactions because the same information is possessed twice, which indeed does not add any value, (ii) removing incomplete data within a dataset which includes corrupted features either due to human error or some other, and (iii) filtering out the features having almost null values. As we know, the missing data need to be dealt with through imputation techniques, but if the feature has more than 90% null values, these imputation techniques will fail miserably.

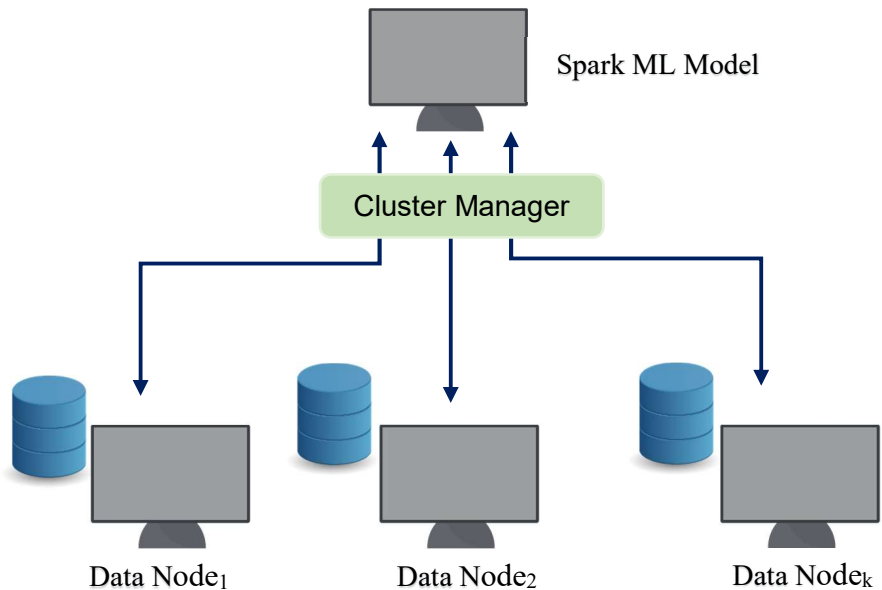

**Fig. 2. Block diagram for ATM fraud detection in static context**

Hence, we dropped off the features having more null values. Thereof, the data can be analyzed to obtain vital insights and check assumptions with the help of generated statistical summary. All the features are normalized. Now, the dataset is observed to be of imbalance type. It is handled by applying various balancing techniques such as ADASYN, SMOTE, etc. The dataset is partitioned into training and test sets in the ratio 80:20. While doing so, stratified random sampling is adopted, where the same proportion of the both positive and negative classes is maintained in both the training and test sets. All of these balancing techniques are discussed in detail in the subsequent sections. On a note, while performing oversampling, the test dataset has to be kept intact. Once, the above operations are performed, the dataset is now stored in RDD (discussed in detail in Section 4.1.1).

Now, the modelling phase begins, where we build the machine learning model such as NB, LR etc. All of these models are built under the spark environment, and we utilized Spark MLlib. This is achieved by constructing a pipeline which consists of a group of stages. Each stage utilizes the data produced by the previous stage. The pipelines in Spark follow Directed Acyclic Graph (DAG), where the stages are specified in topological order. We used feature indexers available in Spark MLlib, to handle the features and then transform the entire data point into a single list. This is required for the Spark MLlib models. Then the model is trained on the training dataset and then tested on the test dataset. The above process is common for the rest of the spark models. While building the models, we used the grid search hyper-parameter tuning technique. On a note, we utilized 10-Fold CV technique and compared the performance thereof. All the metrics area under the receiver operator characteristic curve (AUC), sensitivity and specificity scores are presented and studied effectively.

**Algorithm 1: Sliding Window approach for ATM Fraud detection under streaming context**

*1: Collect the dataset*
*2: Initialize ws ← window size*
*3: Initialize sl ← sliding interval*
*4: Create an Unbounded Table ← φ*
*5: **Start** StreamingContext(Time)*
*6: **while** IncomingStreams == True **do***
*7:      UnboundedTable ← Append the IncomingStream*
*8:      **If** Window == Full **then***
*9:          TrainData  ← Window[:ws-1]*
*10:         TestData ← Window[ws]*
*11:         Train the Model on TrainData*
*12:         Validate the Model on TestData*
*13:         AUC ← Compute AUC score on TestData*
*14:         Sensitivity ← Compute Sensitivity on TestData*
*15:         Specificity ← Compute Specificity on TestData*
*16:         Report AUC, Sensitivity and Sensitivity*
*17:      **end if***
*18:    Slide window by according to sl across UnboundedTable*
*19: **end while***
*20: **Stop** StreamingContext*

## 3.2 Sliding Window for Streaming Data

The methodology discussed in Section 3.1 is very much suitable in the static environment. However, when it comes to processing the stream of data that needs to be handled with the specified interval of time demands a new mechanism. To meet these challenges and demands, we proposed a sliding window inspired streaming approach for ATM fraud data detection. The algorithm is presented in Algorithm 1, and the block diagram is depicted in Fig. 3.

Spark provides DStream (discussed in detail in Section 4.1.2), an array of RDDs where each RDD comprises the transaction information collected at a certain interval of time. All the operations which are to be handled by DStream are processed in a streaming context. Hence, one needs to initialize the streaming context at the beginning of the process and need to close it once the operations are finished. All the variables, models, broadcasting variables etc., are confined to this streaming context. On a note, Spark provides the batch stream processing. Hence, the operations are performed on the data which is collected in the form of batches. By using the ingestion tools, streaming data is processed into the HDFS cluster. In our context, we didn't use any ingestion tool. However, we simulated the data ingestion by using QStreams (discussed in detail in Section 4.1.3) which is provided by Spark. We created an unbounded data frame, where the streams of data are coming in multiple batches.

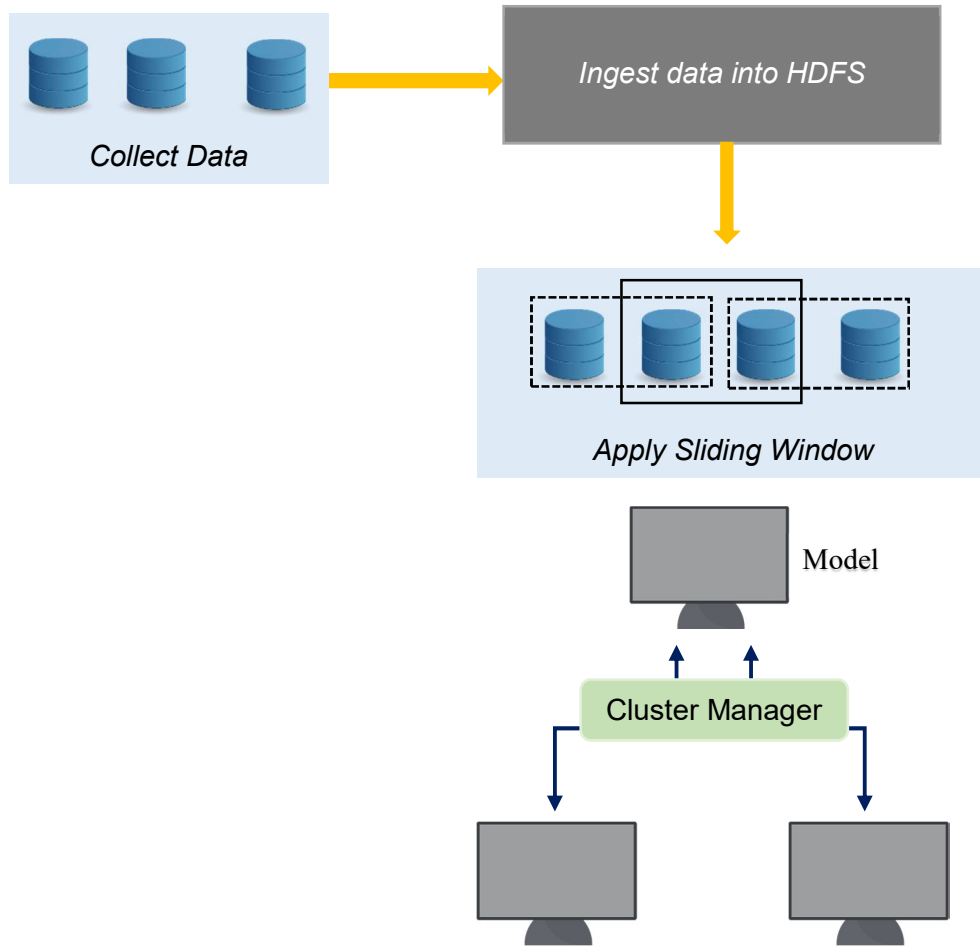

**Fig. 3. Block diagram for ATM fraud detection in streaming context**

In our approach, we have utilized the sliding window approach. The sliding window approach is designed as follows: transactions data streams are collected in the form of batches with the specified time interval. The data collected thus far collected over a particular interval of time is sitting in a single RDD. The number of RDDs collected are dependent on the window size. Too much window size increases the latency. Let the sliding window length be *ws*, and then the training is done on the first '*ws-1*' batches and testing on the latter batch of the sliding window. If the required number of batches has not yet been collected, it will wait until it is full. As discussed earlier, we know that a large window size demands more previous data to be processed. It further increases the latency too. By considering these criteria, we fixed the window size as 2. By doing so, we are restricting ourselves to the most recent data. Hence, the underlying machine learning model is trained on the first RDD of the window and then tested on the second RDD of the window. We kept the stride, the moving speed of the window to be 1. Hence, now the second RDD of the second window becomes the first RDD of the next window. This process is continued until the new data is available and ingested into the cluster.

## 4. Overview of the techniques employed

This section briefly discusses the overall machine learning techniques employed, various balancing techniques, and one-class classification techniques in the current research.

## 4.1. Spark Abstractions

### 4.1.1 RDD

Spark offers the abstraction known as Resilient Distributed Dataset (RDD). It is a collection of dataset partitions that are dispersed throughout the cluster and is immutable. This makes use of Spark to carry out the simultaneous execution. In the event of a failure, it has fault tolerance features that are helpful for retrieving the data. It primarily supports (i) Transformations and (ii) Actions, which are two distinct procedures. Transformations are the set of operations which are used to transform the dataset from one form to other. On each element of RDD, these operations are carried out. Actions are aggregated operations, and hence the result is collected back to the driver. Lazy evaluation is supported by Spark RDD. In order to apply the changes whenever necessary, it first remembers them.

### 4.1.2 Discretized Stream

Fig. 4 shows Discretized Stream (DStream) another abstraction made available by Spark Streaming. It is obvious that there is a constant flow of information in a streaming environement. As a result, DStream contains an ongoing collection of RDDs. Here each RDD consists of the information gathered during a specific period of time. These DStream RDDs allow for the application of all operations, including Transformations and Actions. DStream is an immutable distributed stream that enables parallel operations, just as RDDs. The operations which are applied over DStream is inherently applied over these RDDs.

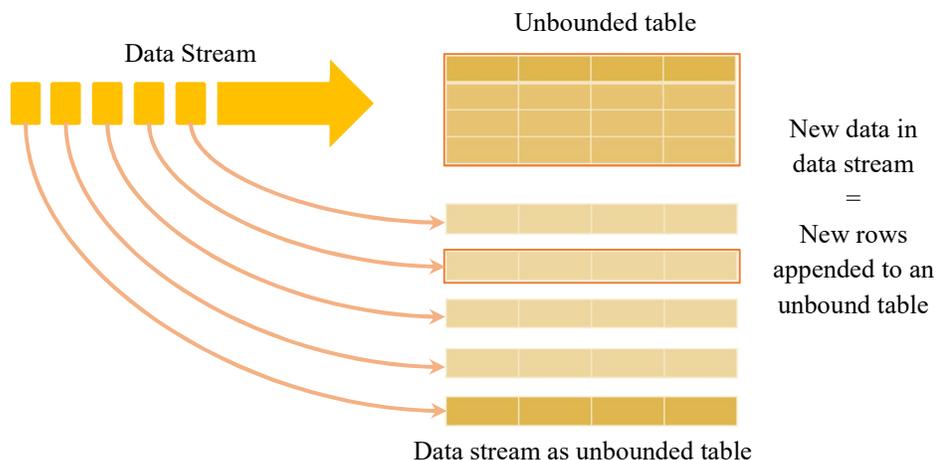

**Fig. 4. DStreams**

### 4.1.3 Queue Streams & Window Operations

In the real world, the dataset is collected over a certain time frame. Such collected streams are processed in batches using Spark. As a result, the underlying queue of RDDs is considered as a batch of RDDs. As a result, all operations are performed on the batch of RDDs. To support this, Spark has an API called Queue Stream that allows users to create DStreams from a collection of RDDs. It is useful in generating an input stream from a queue of RDDs amassed over time. Such generated streams may be handled sequentially or simultaneously.

Additionally, Spark also provides the leverage of applying window operations, which is useful to apply transformations over a sliding window of data. All of the RDDs included in the window are merged, and the corresponding operations are carried out over it, as the window glides over a DStream. On a note, all these window operations are applied based on two important factors which are as follows:

(i)     ***window length***: which defines the duration of the window;
(ii)    ***sliding interval***: the interval at which the window operation is performed.

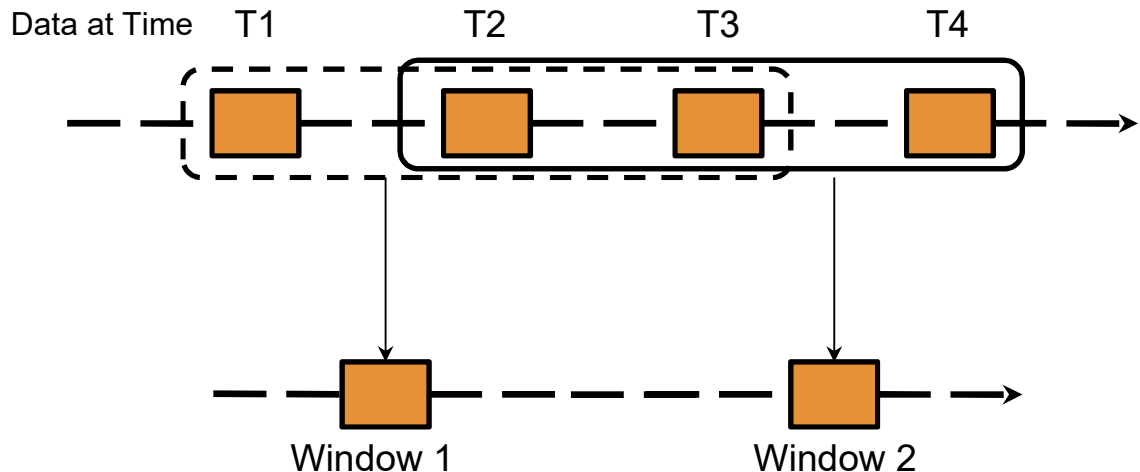

**Fig. 5. Sliding Window Operation on Dstreams (window length size = 3, sliding interval=1)**

## 4.2 Machine Learning techniques

### 4.2.1 Naive Bayes

Naive Bayes (NB) classifiers are the probabilistic models which are based on the Bayes theorem, are used for performing classification tasks. They are fast and easy to implement. NB works under the following assumptions: (i) all the features are independent of each other, and (ii) all the features equally contribute to the decision variable/class variable. These assumptions limit the usage of NB in real-life scenarios, where often one of these assumptions will fail thereby hindering the performance.

### 4.2.2 Logistic Regression

Logistic Regression (LR), is one of the most popular supervised classifiers for the binary classification task. This is also known as the logit model, which estimates the probability of an event occurring such as being fraudulent or legitimate, churned out or non-churn customer etc.

### 4.2.3 Support Vector Machine

Support Vector Machine (SVM), a supervised machine learning algorithm, is proposed by Vapnik [24], is used to tackle various classification and regression tasks. Finding the optimal hyperplane which serves as a decision boundary for classifying the n-dimensional data by maximizing the margin, is the main goal of the SVM. This is formally regarded as structural risk minimization. SVM uses a variety of kernel functions, including linear, radial basis function (RBF), etc.; The kernel functions aid in the transformation of the training data so that a non-linear decision surface may be learned by the SVM.

### 4.2.4 Decision Tree

One of the most effective methods for tackling numerous ML problems including classification and regression is DT [39]. Based on training data, DT creates a tree structure with internal and leaf nodes. Each leaf node represents a class label, while each internal node represents acts as the classification rule. Handling missing values, automated feature selection, handling both categorical and numerical features are a few benefits of DT. The root node splitting is based on various factors, including Gini, entropy etc., and DT generates a set of rules that make it simple to understand. DT is hence regarded as the white box model.

### 4.2.5 Random Forest

RF, proposed by Ho [25], is an ensemble machine-learning technique. Both classification and regression problems can benefit from this. The training phase of RF is carried out as follows: the bootstrap sampling technique is used to choose a random subset of the dataset. One decision tree is constructed independently of the other. The test sample is given to each decision tree during the test phase, and each decision tree may provide a different set of predictions. The majority voting concept is used to make the ultimate decision. Effective handling of the high-dimensional dataset is a hallmark of RF.

### 4.2.6 Gradient Boosting Tree

GBT is also an ensemble machine-learning technique, comprising multiple decision trees. GBT is renowed for handling wide range of classification tasks, and regression tasks. However, the following are where the RF and GBT diverge most: (i) Unlike RF, GBT constructs trees additively, one at a time. Additionally, every new tree fixes the errors made by the ones that came before it. The term "boosting" refers to this method. (ii) The decision is also made earlier rather than later, as it is in RF, by aggregating the tree that has already been created. Hyperparameter tuning must be done extremely carefully because GBT are prone to overfitting.

### 4.2.7 Multi-Layer Perceptron

MLP [26] is one of the most popular neural networks which maps the set of input features to a set of class variables. It is popularly known as a 'universal approximator' in solving both classification and regression problems. MLP consists of three different layers: input, hidden and output layers. The hidden layer tries to identify the non-linear patterns present in the provided dataset. The weights are estimated by using the standard backpropagation algorithm.

## 4.3 Balancing techniques

In this section, we will discuss all the balancing techniques employed in the current study.

### 4.3.1 SMOTE & its variants

**SMOTE**

One of the most popular resampling sampling is SMOTE [35]. It creates new synthesized samples while duplicating the minority class samples. It first chooses a sample at random from the minority class, and then it creates a new synthesized sample by using K-nearest neighbor (KNN). This is repeated until the desired proportion of the samples were generated.

**SMOTE-ENN**

This is developed by Batista et al. [38], and uses the Edited nearest neighbour (ENN) to eliminate some of majority class points and enrich the dataset. SMOTE-ENN is accomplished in two stages, the first of which involves SMOTE generating the samples and the second of which involves ENN removing the samples that go against its guiding principles. According to ENN, a majority class and its K-nearest neighbour are taken out of the dataset if and only if the majority classes of the observations' KNN anyway.

**SMOTE-Tomek**

This is developed by Batista et al. [38], and uses the Tomek links which acts as an undersampling technique to enrich the dataset. Tomek link comprises a set of two points which belongs to two different classes despite being K-nearest neighbour to each other. SMOTE-Tomek is accomplished in two stages, the first of which involves SMOTE generating the samples and the second of which involves Tomek links are removed.

### 4.3.2 ADASYN

The above-discussed SMOTE and its variants don't consider the density of the minority class while generating the synthetic samples. Hence, ADASYN [28] is proposed by incorporating this principle thereby generating a huge number of samples in low regions and fewer / none samples in highly dense regions.

### 4.3.3 Generative Adversarial Networks

GAN was first introduced by Ian Good Fellow et al. [36]. Since then, resampling has been one of several applications where GANs have been utilized. The primary goal of the GAN is to produce data that closely resembles the probabilistic distribution of training data. In this work, we employed two different GAN variants to achieve oversampling in the following way:

**V-GAN**

The main components of GAN [36] are the Discriminator and Generator models, respectively. Here the generator creates the fake samples and tries to deceive the discriminator, while the latter tries to identify the fake samples and categorize them. As a result, the min-max optimization problem is created during GAN training. One at a time, the weights of the generator and discriminator are optimized. Vanilla GAN is another name for this simplistic variant of GAN.

**W-GAN**

Wasserstein distance/ Earth movers distance loss function is the one used by W-GAN [37], which uses a different loss function. The Wasserstein distance is used to compare two distributions for similarity. Both V-GAN and W-GAN follow the same rest of the training process.

## 5 Dataset Description

We collected India-sourced ATM transactional data for the current study. These ATM transactions are tracked by various software that are maintained by Indian banks. This dataset consists of a few days worth of transactions totaling one million transactions over three different channels, including ATM transaction data, point of sale (POS) transaction data, and internet transaction data. This data includes vital facts on this kind of transaction, payment method, client information, transaction amount, etc. We renamed the features in order to preserve data privacy and integrity while adhering to the bank regulations. We employed all the data pre-processing techniques for both static and streaming contexts to clean up the data. After the dataset was refined, it was seen that there were 7,46,724 transactions. The number of features was reduced to 10 as a consequence of the removal of several unique and null-valued characteristics. The data proportion of non-fraud and fraud classes is 87.80:12.20. Hence, it is an imbalanced dataset. All the following experiments are performed in 10-Fold Cross Validation (CV).

### 5.1 Evaluation measures

We considered AUC for measuring the robustness of the employed classifier. In addition to AUC, we also considered sensitivity and specificity as the other measures for the binary classification.

**AUC**

It is proven to be a robust measure while handling imbalanced datasets and is an average of specificity and sensitivity. The mathematical representation of AUC is given in Eq. (2).

$$AUC = \frac{(Sensitivity + Specificity)}{2} \tag{2}$$

**Sensitivity**

It is also well known as True Positive Rate (TPR). It is the ratio of the positive samples that are truly predicted to be positive to all the positive samples. The mathematical notation is given in Eq. (3).

$$Sensitivity = \frac{TP}{TP + FN} \tag{3}$$

where TP is a true positive, and FN is a false negative.

**Specificity**

It is also well known as True Negative Rate (TNR). It is the ratio of the negative samples that are truly predicted to be negative to all the negative samples. The mathematical notation is given in Eq. (4).

$$Specificity = \frac{TN}{TN + FP} \tag{4}$$

where TN is a true negative, and FP is a false positive.

**Table 2: Hyperparameters for all the techniques**

| Model | Hyperparameters |
|-------|-----------------|
| LR | 'regularization parameter': [0.1, 0.01, 0.001]<br>'maximum iterations': [10,20,30,40,50,100,500] |
| SVM | 'regularization parameter': [0.1, 0.01]<br>'maximum iterations': [30,40,50,100] |
| DT | 'criterion': {'gini','entropy'}<br>'maxdepth': [5,10, 15, 20] |
| RF | 'maxdepth': [5, 10, 15, 20]<br>'estimators':[10,20,30,40,50] |
| GBT | 'maxdepth': [5, 10, 15, 20]<br>'estimators':[10,20,30,40,50] |
| MLP | 'solver' : {'l-bfgs', 'gd'}<br>'maximum iterations': [50, 100, 200, 300] |
| V-GAN | **Discriminator:**<br>Hidden layer 1: 128 neurons, activation function:'leaky ReLU'<br>Hidden layer 2:  64 neurons, activation function:'leaky ReLU'<br>Hidden layer 3: 32 neurons, activation function:'leaky ReLU'<br>Hidden layer 4: 8 neurons, activation function:'leaky ReLU'<br>Epochs=10,000 |
| W-GAN | **Discriminator:**<br>Hidden layer 1: 256 neurons, activation function:'leaky ReLU'<br>Hidden layer 2:  128 neurons, activation function:'leaky ReLU'<br>Hidden layer 3: 64 neurons, activation function:'leaky ReLU'<br>Hidden layer 4: 32 neurons, activation function:'leaky ReLU'<br>Epochs=10,000 |

# 6 Results and discussion

In this section, we presented the results in two different contexts viz., static and streaming contexts respectively. Firstly, we will discuss static context results followed by streaming context results.

## 6.1 Static Context results

In the current research study, we utilized the hyperparameters presented in Table 2. We presented all the hyperparameters after performing rigorous experiments. As discussed in Section 3, all the experiments followed the same identical data cleansing, and data pre-processing technique. We observed that the ATM transactions dataset collected from India is highly imbalanced i.e., 87.80 : 12.20. Hence, we incorporated various balancing techniques viz., SMOTE, SMOTE-ENN, V-GAN, etc. It is to be noted that all the experiments are performed in 10-Fold Cross Validation (CV). In the imbalanced classification datasets, AUC is proven to be a robust measure which is quite evident in previous case studies such as credit card fraud detection. Hence, while choosing the best model, AUC is given more preference. The corresponding results are presented in Table 3. Further, we also discussed sensitivity and specificity scores in Table 4 and Table 5 respectively.

Out of all ML models, MLP turned out to be the best model in terms of AUC (refer to Table 3) after balancing the dataset with the V-GAN balancing technique. The second and third place is secured by GBT and RF respectively. It is interesting to note that both RF and GBT obtained similar AUC. The same is also reflected in sensitivity and specificity scores. As a decision maker, while choosing the best model the complexity of the model should also need to be considered along with the numerical AUC. In such a case, RF turned out to be the best model because of its less number of learnable parameters than MLP and low complexity than GBT. Further, it is observed to be having a very minor difference (AUC of 0.2%) in terms of mean AUC when compared to the other two best models. Overall, all the tree-based models performed relatively well when compared to other ML models. In terms of balancing techniques, V-GAN clearly outperformed the other balancing techniques in obtaining the best AUC.

There is another tie-breaker which can be considered while choosing the best model. It is a well-known fact that DT is the explainable model in the tree-based models. RF and GBT are ensemble models. Hence, there are complex models that stand second in terms of explainable aspect. But if one closely observed that the AUC is 2% less than the RF and GBT which is quite not acceptable in critical applications like fraud detection. Hence, despite having lesser interpretability, RF and GBT are ranked over DT. Because of its numerical superiority and less model complexity, RF ranked as the best model.

**Table 3: Mean AUC obtained by various ML models under 10-Fold CV**

| Model | Balancing Technique | | | | | | |
|---|---|---|---|---|---|---|---|
| | Imbalanced | SMOTE | SMOTE-Tomek | SMOTE-ENN | ADASYN | V-GAN | W-GAN |
| NB | 0.721 | 0.775 | 0.775 | 0.777 | 0.692 | 0.897 | 0.798 |
| LR | 0.709 | 0.803 | 0.803 | 0.805 | 0.690 | 0.893 | 0.869 |
| SVM | 0.807 | 0.806 | 0.805 | 0.807 | 0.694 | 0.890 | 0.853 |
| DT | 0.755 | 0.823 | 0.823 | 0.824 | 0.775 | 0.953 | 0.904 |
| RF | 0.759 | 0.837 | 0.837 | 0.838 | 0.791 | **0.975** | 0.920 |
| GBT | 0.759 | 0.837 | 0.837 | 0.838 | 0.791 | 0.975 | 0.920 |
| MLP | 0.757 | 0.837 | 0.831 | 0.831 | 0.721 | **0.977** | 0.904 |

Similar observations were noticed with respect to the sensitivity scores (refer to Table 4). As we know that sensitivity is also another critical parameter that artefacts the number of fraudulent transactions truly identified to be as fraudulent ones. MLP outperformed the rest of the model by obtaining higher mean sensitivity scores. However, when considering the model complexity and numerical superiority as tie-breakers, RF turned out to be the best model in terms of mean sensitivity.

On the other hand, a similar observation is made on the specificity score (refer to Table 5) which artefacts the number of non-fraudulent transactions truly identified to be as non-fraudulent ones. Specificity is also another critical factor because no loyal customer's transaction should be tagged as a fraudulent transaction. It is observed that SVM obtained a sensitivity score of 1.0 yet the specificity score is remarkably inferior to tree-based models. After considering tie-breakers, we concluded that RF turned out to be the best model in the static context.

**Table 4: Mean Sensitivity scores obtained by various ML techniques on 10-Fold CV**

| Model | Balancing Technique | | | | | | |
|---|---|---|---|---|---|---|---|
| | Imbalanced | SMOTE | SMOTE-Tomek | SMOTE-ENN | ADASYN | V-GAN | W-GAN |
| NB | 0.546 | 0.779 | 0.778 | 0.780 | 0.699 | 0.871 | 0.674 |
| LR | 0.540 | 0.782 | 0.781 | 0.783 | 0.660 | 0.787 | 0.745 |
| SVM | 0.736 | 0.775 | 0.774 | 0.775 | 0.632 | 0.781 | 0.707 |
| DT | 0.607 | 0.789 | 0.787 | 0.789 | 0.738 | 0.930 | 0.824 |
| RF | 0.618 | 0.773 | 0.773 | 0.776 | 0.769 | 0.961 | 0.850 |
| GBT | 0.618 | 0.773 | 0.773 | 0.776 | 0.769 | 0.961 | 0.850 |
| MLP | 0.613 | 0.781 | 0.781 | 0.781 | 0.675 | **0.969** | 0.825 |

**Table 5: Mean Specificity scores obtained by various ML techniques on 10-Fold CV**

| Model | Balancing Technique | | | | | | |
|---|---|---|---|---|---|---|---|
| | Imbalanced | SMOTE | SMOTE-Tomek | SMOTE-ENN | ADASYN | V-GAN | W-GAN |
| NB | 0.895 | 0.772 | 0.773 | 0.774 | 0.685 | 0.924 | 0.922 |
| LR | 0.878 | 0.825 | 0.825 | 0.826 | 0.720 | 0.999 | 0.992 |
| SVM | 0.877 | 0.837 | 0.836 | 0.838 | 0.757 | **1.0** | 0.999 |
| DT | 0.902 | 0.857 | 0.859 | 0.859 | 0.813 | 0.975 | 0.991 |
| RF | 0.901 | 0.901 | 0.902 | 0.901 | 0.812 | 0.989 | 0.991 |
| GBT | 0.901 | 0.901 | 0.902 | 0.901 | 0.812 | 0.989 | 0.991 |
| MLP | 0.902 | 0.882 | 0.882 | 0.882 | 0.767 | 0.986 | 0.982 |

## 6.2 Statistical testing of the results

The two-tailed t-test analysis is conducted to check whether model A's numerical superiority over model B is purely a coincidence or, indeed, due to its superior nature. Hence, we conducted a pairwise t-test analysis with the best performing model to the rest. The t-test analysis is conducted on the AUC obtained by models over 10 fold cross validation which are reported in Table 3. The t-test results are reported in Table 6.

The null hypothesis is, *$H_0$: both the algorithms are statistically equal*.
The alternate hypothesis is, *$H_1$: both the algorithms are statistically not equal*.

The p-value determines whether to accept or reject the null hypothesis. The significance level is chosen to be 5%, and the degree of freedom is 18 (10+10-2). Hence, the p-value should be less than 5% to reject the null hypothesis. Table 6 infers that when the p-values are less than 0.05; hence the null hypothesis is rejected, and the alternate hypothesis is accepted in the t-test conducted between RF and DT. This concludes that in the static context, RF is more statistically significant than DT in terms of AUC. However, RF and GBT are proven to be statistically equal. However, in terms of complexity RF is less than GBT which makes RF to be considered as the best model.

**Table 6: Paired t-test results**

| Model | Parameter | |
|---|---|---|
| | t-statistic | p-value |
| RF vs GBT | 0.0 | 1.0 |
| RF vs DT | 33.39 | $1.21 \times 10^{-17}$ |

### 6.3 Streaming Context results

Now, we will discuss the performance of the models in the streaming context. As we already discussed, the ATM transaction dataset is highly imbalanced. Hence, the transactions collected over a window might not have any fraudulent transaction. Further, the binary classification models cannot be built when only one single class-related data point are given. Hence, in our approach, one assumption is made in the streaming context with respect to proportion of the negative and positive class transactions collected over a window. Hence, in every data stream, we will include the fraudulent transactions (positive class samples) with the collected ATM transactions. Hence, the balanced datasets are only considered for being evaluated in the streaming context. Further, latency is another critical parameter while evaluating the model in the streaming context. Hence, the models should be less complex. Therefore, we considered the simple and tree-based models and compared the performance. It is important to note that all the models are evaluated in a sliding window fashion, and the metrics such as AUC, sensitivity, and specificity are calculated and discussed in this section. After employing the balancing technique, the dataset has now become 10 lakh transactions. Each window comprises 1000 transactions thereby making the number of windows as 1000.

**Table 7: Mean AUC scores obtained by various ML techniques over sliding windows**

| Model | Balancing Technique | | | | | |
|---|---|---|---|---|---|---|
| | SMOTE | SMOTE-Tomek | SMOTE-ENN | ADASYN | V-GAN | W-GAN |
| LR | 0.811 | 0.811 | 0.811 | 0.691 | 0.842 | 0.875 |
| KNN | 0.854 | 0.855 | 0.855 | 0.774 | 0.870 | 0.896 |
| DT | 0.889 | 0.889 | 0.889 | 0.810 | 0.908 | 0.909 |
| RF | 0.896 | 0.898 | 0.898 | 0.821 | **0.910** | 0.911 |

Further, here also, we considered AUC as the suitable metric to choose the best model, and the results are presented in Table 7. We presented AUC obtained in Fig. 6 to Fig. 11 to illustrate the outlier behaviour of the employed ML models. We further presented AUC obtained over different windows in Fig. 12 to Fig. 15 to give an abstract of the performance of various models. This will indeed give a depiction over the performance of ML model. Overall, RF model performed the best irrespective of the balancing technique. DT stands second on the list. Further, Fig. 10 shows that RF model has very fewer outlier behaviour which makes it more stable while identifying the positive class. However, DT showed higher variability which makes it be second despite having interpretability aspect. The same is observed in the sensitivity and specificity scores. Interestingly, RF not only showed superior performance in mean AUC, sensitivity and specificity scores but also the showed lesser variability which supports its robustness. The same observation is noticed with respect to other balancing techniques (refer to Fig. 6 to Fig. 11). V-GAN is empirically obtained to be an efficient balancing technique in this setup which is followed by SMOTE-ENN.

**Table 8: Mean Sensitivity scores obtained by various ML techniques over sliding windows**

| Model | Balancing Technique | | | | | |
| | SMOTE | SMOTE-Tomek | SMOTE-ENN | ADASYN | V-GAN | W-GAN |
|---|---|---|---|---|---|---|
| **LR** | 0.806 | 0.806 | 0.806 | 0.667 | 0.848 | 0.914 |
| **KNN** | 0.860 | 0.861 | 0.861 | 0.746 | 0.828 | 0.859 |
| **DT** | 0.895 | 0.895 | 0.895 | 0.779 | 0.865 | 0.867 |
| **RF** | 0.899 | 0.895 | 0.895 | 0.793 | **0.871** | 0.873 |

**Table 9: Mean Specificity scores obtained by various ML techniques over sliding windows**

| Model | Balancing Technique | | | | | |
| | SMOTE | SMOTE-Tomek | SMOTE-ENN | ADASYN | V-GAN | W-GAN |
|---|---|---|---|---|---|---|
| **LR** | 0.815 | 0.816 | 0.816 | 0.715 | 0.836 | 0.836 |
| **KNN** | 0.849 | 0.850 | 0.850 | 0.801 | 0.913 | 0.933 |
| **DT** | 0.882 | 0.882 | 0.882 | 0.842 | **0.950** | 0.952 |
| **RF** | 0.893 | 0.892 | 0.892 | 0.810 | 0.948 | 0.949 |

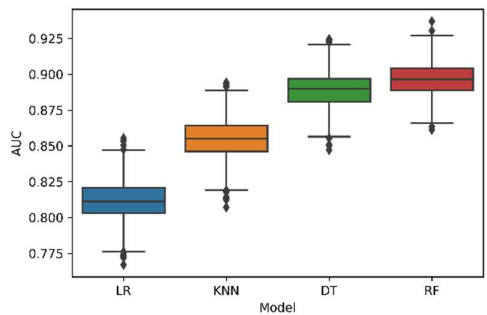 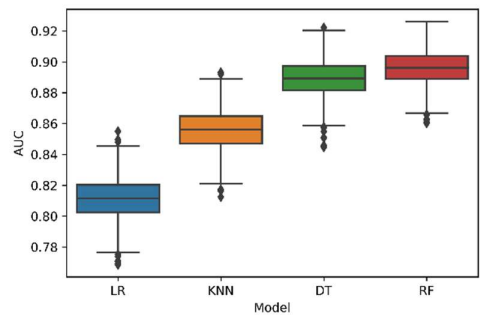

**Fig. 6. AUC obtained when SMOTE is employed in the streaming context**      **Fig. 7. AUC obtained when SMOTE-Tomek is employed in the streaming context**

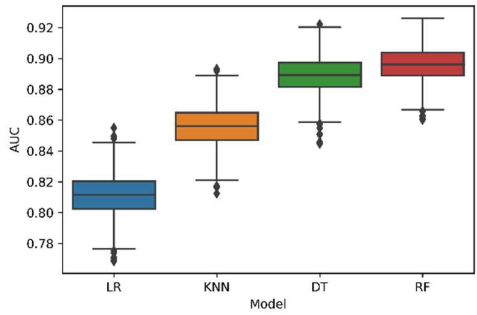

**Fig. 8.** AUC obtained when SMOTE-ENN is employed in the streaming context

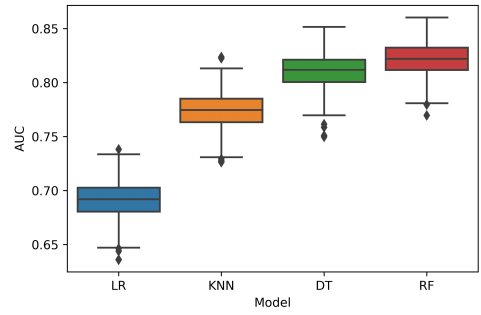

**Fig. 9.** AUC obtained when ADASYN is employed in the streaming context

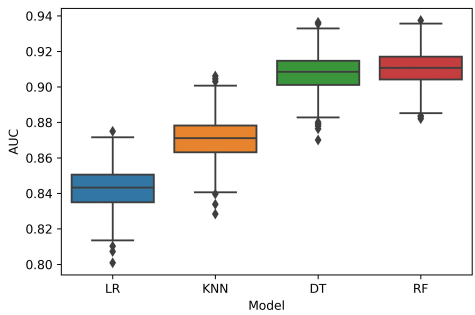

**Fig. 10.** AUC obtained when V-GAN is employed in the streaming context

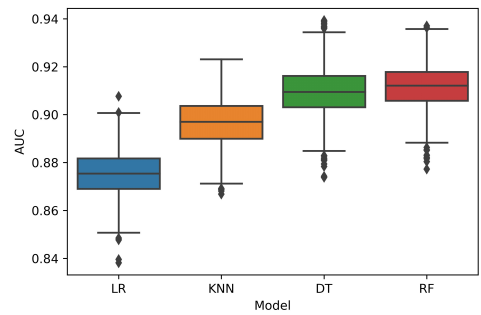

**Fig. 11.** AUC obtained when W-GAN is employed in the streaming context

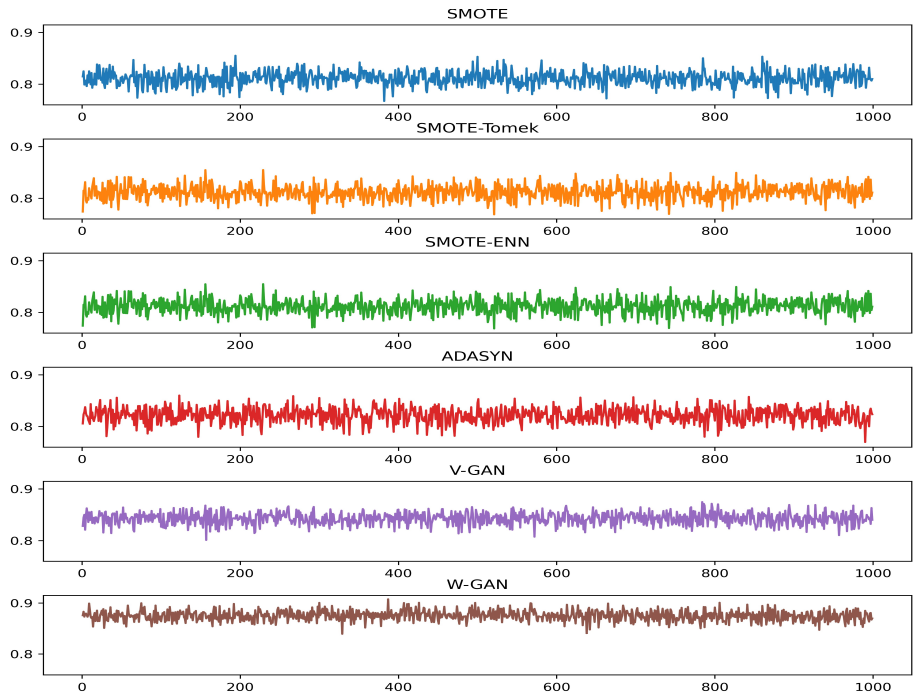

**Fig. 12.** AUC obtained by LR model over several windows

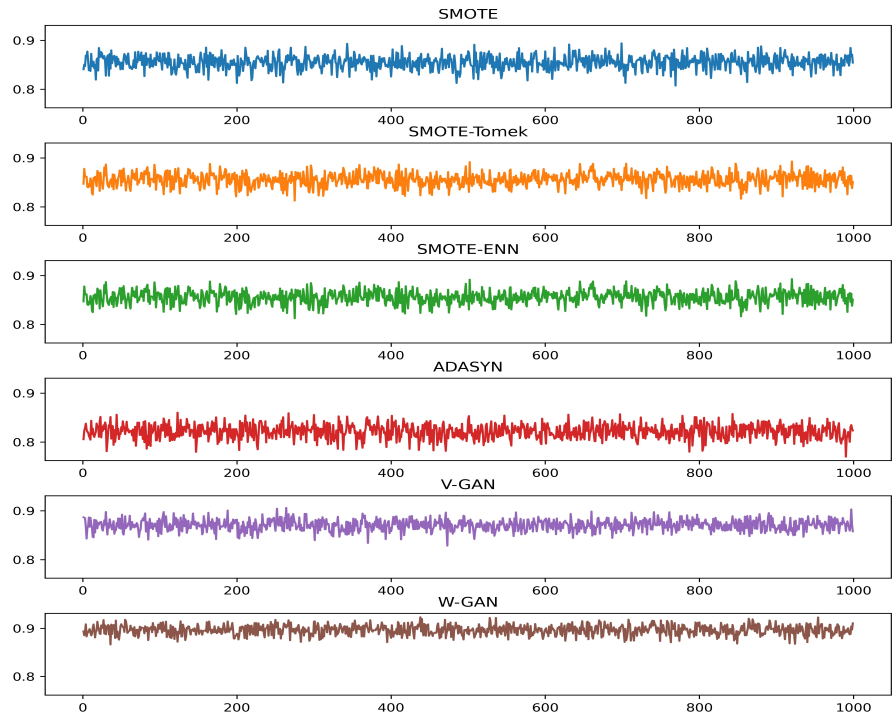

**Fig. 13. AUC obtained by KNN model over several windows**

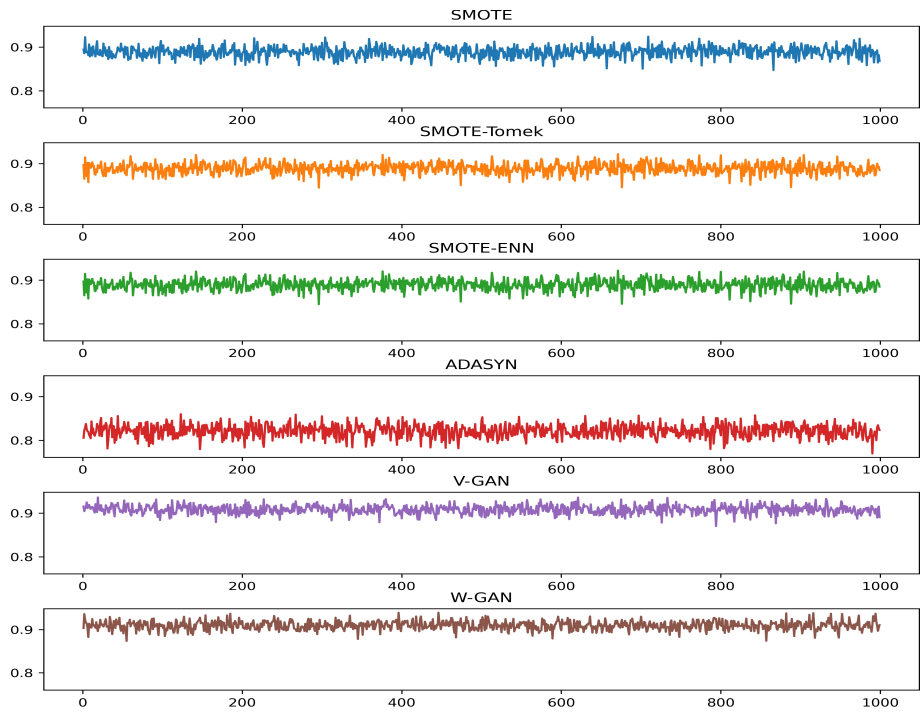

**Fig. 14. AUC obtained by DT model over several windows**

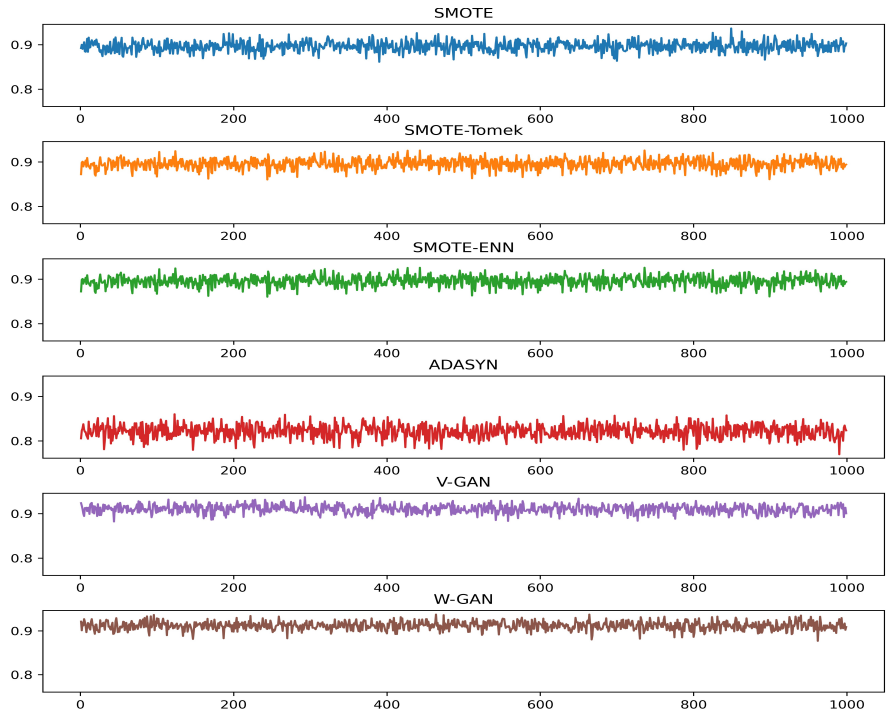

**Fig. 15. AUC obtained by RF model over several windows**

### 6.4 Statistical testing of the results

The Wilcoxon-test is a non-parametric test is performed to check whether the superior performance of model A over model B is purely a coincidence or due to its true superior nature. Hence, we conducted Wilcoxon test analysis on top-2 models by considering AUC obtained over several windows. The Wilcoxon test results are presented in Table 10.

The null hypothesis is, $H_0$: *both the algorithms are statistically equal*.
The alternate hypothesis is, $H_1$: *both the algorithms are statistically not equal*.

The p-value determines whether to accept or reject the null hypothesis. The significance level is chosen to be 5%. Hence, the p-value should be less than 5% to reject the null hypothesis. Table 10 infers that in the streaming context, RF is more statistically significant than the next-best performing model DT.

**Table 10: Paired Wilcoxon-test results**

| Model | Parameter | |
|---|---|---|
| | Wilcoxon test-statistic | p-value |
| DT vs RF | 204017.0 | $5.34 \times 10^{-07}$ |

## 7. Conclusions and Limitations

In the current study, we developed a scalable ML models for the identification of ATM fraudulent transactions in the ATM transaction dataset collected from India. We devised two different mechanisms suitable for both streaming and static environments respectively. The collected dataset is observed to be highly imbalanced. Therefore, we incorporated various sampling techniques viz., SMOTE,

ADASYN, and GAN. Of all the sampling techniques, it turned out that V-GAN significantly outperformed the rest of the models. Further, we employed various machine learning techniques in the static and streaming contexts. In the static context, RF outperformed the rest of the models by achieving AUC of 0.973 followed by GBT with similar AUC. In the streaming context, RF secured first place. Further, we conducted statistical test which turned out to be statistically significant.

Our proposed sliding window approach helps to mitigate the risk of correctly identifying fraudulent ATM transactions. However, we observed that there is one certain limitation which is as follows: every window assumes that it has received a certain amount of positive sample information. To mitigate this risk, while training the model, we need to provide some historic positive samples along with the collected data at the same timestamp. Otherwise, any ML technique fails to correctly identify the fraudulent samples. Alternatively, we can employ One Class Classification (OCC) model instead of binary classification models.